# SPEECH DIALOGUE WITH FACIAL DISPLAYS: MULTIMODAL HUMAN-COMPUTER CONVERSATION

**Katashi Nagao** and **Akikazu Takeuchi**

Sony Computer Science Laboratory Inc.
3–14–13 Higashi-gotanda, Shinagawa–ku, Tokyo 141, Japan
E-mail: {nagao,takeuchi}@csl.sony.co.jp

## Abstract

Human face-to-face conversation is an ideal model for human-computer dialogue. One of the major features of face-to-face communication is its multiplicity of communication channels that act on multiple modalities. To realize a natural multimodal dialogue, it is necessary to study how humans perceive information and determine the information to which humans are sensitive. A face is an independent communication channel that conveys emotional and conversational signals, encoded as facial expressions. We have developed an experimental system that integrates speech dialogue and facial animation, to investigate the effect of introducing communicative facial expressions as a new modality in human-computer conversation. Our experiments have shown that facial expressions are helpful, especially upon first contact with the system. We have also discovered that featuring facial expressions at an early stage improves subsequent interaction.

## Introduction

Human face-to-face conversation is an ideal model for human-computer dialogue. One of the major features of face-to-face communication is its multiplicity of communication channels that act on multiple modalities. A channel is a communication medium associated with a particular encoding method. Examples are the auditory channel (carrying speech) and the visual channel (carrying facial expressions). A modality is the sense used to perceive signals from the outside world.

Many researchers have been developing multimodal dialogue systems. In some cases, researchers have shown that information in one channel complements or modifies information in another. As a simple example, the phrase "delete it" involves the coordination of voice with gesture. Neither makes sense without the other. Researchers have also noticed that nonverbal (gesture or gaze) information plays a role in setting the situational context which is useful in restricting the hypothesis space constructed during language processing. Anthropomorphic interfaces present another approach to multimodal dialogues. An anthropomorphic interface, such as Guides (Don *et al.*, 1991), provides a means to realize a new style of interaction. Such research attempts to computationally capture the communicative power of the human face and apply it to human-computer dialogue.

Our research is closely related to the last approach. The aim of this research is to improve human-computer dialogue by introducing human-like behavior into a speech dialogue system. Such behavior will include factors such as facial expressions and head and eye movement. It will help to reduce any stress experienced by users of computing systems, lowering the complexity associated with understanding system status.

Like most dialogue systems developed by natural language researchers, our current system can handle domain-dependent, information-seeking dialogues. Of course, the system encounters problems with ambiguity and missing information (i.e., anaphora and ellipsis). The system tries to resolve them using techniques from natural language understanding (e.g., constraint-based, case-based, and plan-based methods). We are also studying the use of synergic multimodality to resolve linguistic problems, as in conventional multimodal systems. This work will be reported in a separate publication.

In this paper, we concentrate on the role of nonverbal modality for increasing flexibility of human-computer dialogue and reducing the mental barriers that many users associate with computer systems.

## Research Overview of Multimodal Dialogues

Multimodal dialogues that combine verbal and nonverbal communication have been pursued mainly from the following three viewpoints.

1. *Combining direct manipulation with natural language (deictic) expressions*

   "Direct manipulation (DM)" was suggested by Shneiderman (1983). The user can interact directly with graphical objects displayed on the computer screen with rapid, incremental, reversible operations whose effects on the objects of interest are immediately visible.

   The semantics of natural language (NL) expressions is anchored to real-world objects and events by means of pointing and demonstrating actions and deictic expressions such as "this," "that," "here," "there," "then," and "now." Some research on dialogue systems has combined deictic gestures and natural language such as Put-That-There (Bolt, 1980), CUBRICON (Neal *et al.*, 1988), and ALFRESCO (Stock, 1991).

   One of the advantages of combined NL/DM interaction is that it can easily resolve the missing information in NL expressions. For example, when the system receives a user request in speech like "delete that object," it can fill in the missing information by looking for a pointing gesture from the user or objects on the screen at the time the request is made.

2. *Using nonverbal inputs to specify the context and filter out unrelated information*

   The focus of attention or the focal point plays a very important role in processing applications with a broad hypothesis space such as speech recognition. One example of focusing modality is following the user's looking behavior. Fixation or gaze is useful for the dialogue system to determine the context of the user's interest. For example, when a user is looking at a car, that the user says at that time may be related to the car. Prosodic information (e.g., voice tones) in the user's utterance also helps to determine focus. In this case, the system uses prosodic information to infer the user's beliefs or intentions. Combining gestural information with spoken language comprehension shows another example of how context may be determined by the user's nonverbal behavior (Oviatt *et al.*, 1993). This research uses multimodal forms that prompt a user to speak or write into labeled fields. The forms are capable of guiding and segmenting inputs, of conveying the kind of information the system is expecting, and of reducing ambiguities in utterances by restricting syntactic and semantic complexities.

3. *Incorporating human-like behavior into dialogue systems to reduce operation complexity and stress often associated with computer systems*

   Designing human-computer dialogue requires that the computer makes appropriate backchannel feedbacks like nodding or expressions such as "aha" and "I see." One of the major advantages of using such nonverbal behavior in human-computer conversation is that reactions are quicker than those from voice-based responses. For example, the facial backchannel plays an important role in human face-to-face conversation. We consider such quick reactions as being situated actions (Suchman, 1987) which are necessary for resource-bounded dialogue participants. Timely responses are crucial to successful conversation, since some delay in reactions can imply specific meanings or make messages unnecessarily ambiguous.

   Generally, visual channels contribute to quick user recognition of system status. For example, the system's gaze behavior (head and eye movement) gives a strong impression of whether it is paying attention or not. If the system's eyes wander around aimlessly, the user easily recognizes the system's attention elsewhere, perhaps even unaware that he or she is speaking to it. Thus, gaze is an important indicator of system (in this case, speech recognition) status.

   By using human-like nonverbal behavior, the system can more flexibly respond to the user than is possible by using verbal modality alone.

We focused on the third viewpoint and developed a system that acts like a human. We employed communicative facial expressions as a new modality in human-computer conversation. We have already discussed this, however, in another paper (Takeuchi and Nagao, 1993). Here, we consider our implemented system as a testbed for incorporating human-like (nonverbal) behavior into dialogue systems.

The following sections give a system overview, an example dialogue along with a brief explanation of the process, and our experimental results.

## Incorporating Facial Displays into a Speech Dialogue System

### Facial Displays as a New Modality

The study of facial expressions has attracted the interest of a number of different disciplines, including psychology, ethology, and interpersonal communications. Currently, there are two basic schools of thought. One regards facial expressions as being expressions of emotion (Ekman and Friesen, 1984). The other views facial expressions in a social context, regarding them as being communicative signals (Chovil, 1991). The term "facial displays" is essentially the same as "facial

expressions," but is less reminiscent of emotion. In this paper, therefore, we use "facial displays."

A face is an independent communication channel that conveys emotional and conversational signals, encoded as facial displays. Facial displays can be also regarded as being a modality because the human brain has a special circuit dedicated to their processing.

Table 1 lists all the communicative facial displays used in the experiments described in a later section. The categorization framework, terminology, and individual displays are based on the work of Chovil (1991), with the exception of the emphasizer, underliner, and facial shrug. These were coined by Ekman and Friesen (1969).

Table 1: Communicative Facial Displays Used in the Experiments. (Categorization based mostly on Chovil [1991])

| Syntactic Display | |
|---|---|
| 1. Exclamation mark | Eyebrow raising |
| 2. Question mark | Eyebrow raising or lowering |
| 3. Emphasizer | Eyebrow raising or lowering |
| 4. Underliner | Longer eyebrow raising |
| 5. Punctuation | Eyebrow movement |
| 6. End of an utterance | Eyebrow raising |
| 7. Beginning of a story | Eyebrow raising |
| 8. Story continuation | Avoid eye contact |
| 9. End of a story | Eye contact |
| Speaker Display | |
| 10. Thinking/Remembering | Eyebrow raising or lowering, closing the eyes, pulling back one mouth side |
| 11. Facial shrug: "I don't know" | Eyebrow flashes, mouth corners pulled down, mouth corners pulled back |
| 12. Interactive: "You know?" | Eyebrow raising |
| 13. Metacommunicative: Indication of sarcasm or joke | Eyebrow raising and looking up and off |
| 14. "Yes" | Eyebrow actions |
| 15. "No" | Eyebrow actions |
| 15. "Not" | Eyebrow actions |
| 17. "But" | Eyebrow actions |
| Listener Comment Display | |
| 18. Backchannel: Indication of attendance | Eyebrow raising, mouth corners turned down |
| 19. Indication of loudness | Eyebrows drawn to center |
| Understanding levels | |
| 20. Confident | Eyebrow raising, head nod |
| 21. Moderately confident | Eyebrow raising |
| 22. Not confident | Eyebrow lowering |
| 23. "Yes" | Eyebrow raising |
| Evaluation of utterances | |
| 24. Agreement | Eyebrow raising |
| 25. Request for more information | Eyebrow raising |
| 26. Incredulity | Longer eyebrow raising |

Three major categories are defined as follows.

**Syntactic displays.** These are facial displays that (1) place stress on particular words or clauses, (2) are connected with the syntactic aspects of an utterance, or (3) are connected with the organization of the talk.

**Speaker displays.** Speaker displays are facial displays that (1) illustrate the idea being verbally conveyed, or (2) add additional information to the ongoing verbal content.

**Listener comment displays.** These are facial displays made by the person who is not speaking, in response to the utterances of the speaker.

## An Integrated System of Speech Dialogue and Facial Animation

We have developed an experimental system that integrates speech dialogue and facial animation to investigate the effects of human-like behavior in human-computer dialogue.

The system consists of two subsystems, a facial animation subsystem that generates a three-dimensional face capable of a range of facial displays, and a speech dialogue subsystem that recognizes and interprets speech, and generates voice outputs. Currently, the animation subsystem runs on an SGI 320VGX and the speech dialogue subsystem on a Sony NEWS workstation. These two subsystems communicate with each other via an Ethernet network.

Figure 1 shows the configuration of the integrated system. Figure 2 illustrates the interaction of a user with the system.

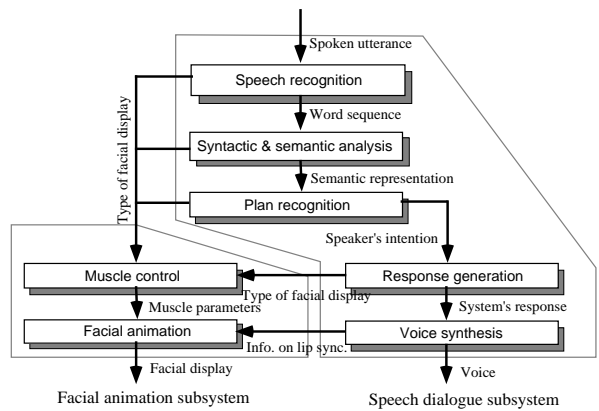

Figure 1: System Configuration

### Facial Animation Subsystem

The face is modeled three-dimensionally. Our current version is composed of approximately 500 polygons. The face can be rendered with a skin-like surface material, by applying a texture map taken from a photograph or a video frame.

In 3D computer graphics, a facial display is realized by local deformation of the polygons representing the face. Waters showed that deformation that simulates the action of muscles underly-ing the face looks more natural (Waters, 1987). We therefore use numerical equations to simulate muscle actions, as defined by Waters. Currently, the system incorporates 16 muscles and 10 parameters, controlling mouth opening, jaw rotation, eye movement, eyelid opening, and head orientation. These 16 muscles were determined by Waters, considering the correspondence with action units in the Facial Action Coding System (FACS) (Ekman and Friesen, 1978). For details of the facial modeling and animation system, see (Takeuchi and Franks, 1992).

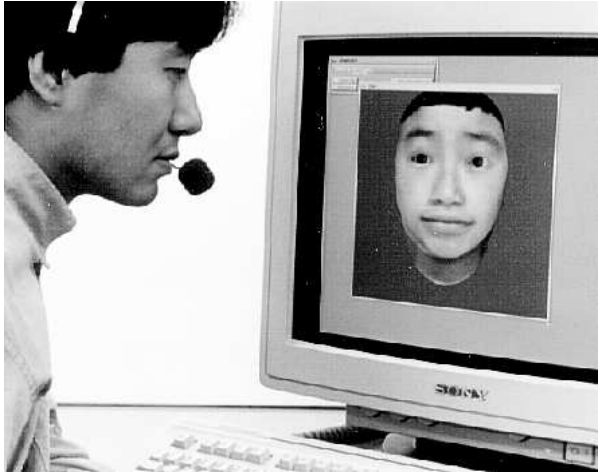

Figure 2: Dialogue Snapshot

We use 26 synthesized facial displays, corresponding to those listed in Table 1, and two additional displays. All facial displays are generated by the above method, and rendered with a texture map of a young boy's face. The added displays are "Smile" and "Neutral." The "Neutral" display features no muscle contraction whatsoever, and is used when no conversational signal is needed.

At run-time, the animation subsystem awaits a request from the speech subsystem. When the animation subsystem receives a request that specifies values for the 26 parameters, it starts to deform the face, on the basis of the received values. The deformation process is controlled by the differential equation $f' = a - f$, where $f$ is a parameter value at time $t$ and $f'$ is its time derivative at time $t$. $a$ is the target value specified in the request. A feature of this equation is that deformation is fast in the early phase but soon slows, corresponding closely to the real dynamics of facial displays. Currently, the base performance of the animation subsystem is around 20–25 frames per second when running on an SGI Power Series. This is sufficient to enable real-time animation.

## Speech Dialogue Subsystem

Our speech dialogue subsystem works as follows. First, a voice input is acoustically analyzed by a built-in sound processing board. Then, a speech recognition module is invoked to output word sequences that have been assigned higher scores by a probabilistic phoneme model. These word sequences are syntactically and semantically analyzed and disambiguated by applying a relatively loose grammar and a restricted domain knowledge. Using a semantic representation of the input utterance, a plan recognition module extracts the speaker's intention. For example, from the utterance "I am interested in Sony's workstation," the module interprets the speaker's intention as "he wants to get precise information about Sony's workstation." Once the system determines the speaker's intention, a response generation module is invoked. This generates a response to satisfy the speaker's request. Finally, the system's response is output as voice by a voice synthesis module. This module also sends the information about lip synchronization that describes phonemes (including silence) in the response and their time durations to the facial animation subsystem.

With the exception of the voice synthesis module, each module can send messages to the facial animation subsystem to request the generation of a facial display. The relation between the speech dialogues and facial displays is discussed later.

In this case, the specific task of the system is to provide information about Sony's computer-related products. For example, the system can answer questions about price, size, weight, and specifications of Sony's workstations and PCs.

Below, we describe the modules of the speech dialogue subsystem.

**Speech recognition.** This module was jointly developed with the Electrotechnical Laboratory and Tokyo Institute of Technology. Speaker-independent continuous speech inputs are accepted without special hardware. To obtain a high level of accuracy, context-dependent phonetic hidden Markov models are used to construct phoneme-level hypotheses (Itou et al., 1992). This module can generate N-best word-level hypotheses.

**Syntactic and semantic analysis.** This module consists of a parsing mechanism, a semantic analyzer, a relatively loose grammar consisting of 24 rules, a lexicon that includes 34 nouns, 8 verbs, 4 adjectives and 22 particles, and a frame-based knowledge base consisting of 61 conceptual frames. Our semantic analyzer can handle ambiguities in syntactic structures and generates a semantic representation of the speaker's utterance. We applied a preferential constraint satisfaction technique (Nagao, 1992) for performing disambiguation and semantic analysis. By allowing the preferences to control the application of the constraints, ambiguities can be efficiently resolved, thus avoiding combinatorial explosions.

**Plan recognition.** This module determines the speaker's intention by constructing a model of his/her beliefs, dynamically adjusting and expanding the model as the dialogue progresses (Nagao, 1993). The model deals with the dynamic nature of dialogues by applying the following two mechanisms. First, preferences among the contexts are dynamically computed based on the facts and assumptions within each context. The preference provides a measure of the plausibility of a context. The currently most preferable context contains a currently recognized plan. Secondly, changing the most plausible context among mutually exclusive contexts within a dialogue is formally treated as belief revision of a plan-recognizing agent. However, in some dialogues, many alternatives may have very similar preference values. In this situation, one may wish to obtain additional information, allowing one to be more certain about committing to the preferable context. A criterion for detecting such a critical situation based on the preference measures for mutually exclusive contexts is being explored. The module also maintains the topic of the current dialogue and can handle anaphora (reference of pronouns) and ellipsis (omission of subjects).

**Response generation.** This module generates a response by using domain knowledge (database) and text templates (typical patterns of utterances). It selects appropriate templates and combines them to construct a response that satisfies the speaker's request.

In our prototype system, the method used to comprehend speech is a specific combination of specific types of knowledge sources with a rather fixed information flow, preventing flexible interaction between them. A new method that enables flexible control of omni-directional information flow in a very context-sensitive fashion has been announced (Nagao et al., 1993). Its architecture is based on *dynamical constraint* (Hasida et al., 1993) which defines a fine classification based on the dimensions of satisfaction and the violation of constraints. A constraint is represented in terms of a clausal logic program. A fine-grained declarative semantics is defined for this constraint by measuring the degree of violation in terms of real-valued *potential energy*. A field of force arises along the gradient of this energy, inferences being controlled on the basis of the dynamics. This allows us to design combinatorial behaviors under declarative semantics within tractable computational complexity. Our forthcoming system can, therefore, concentrate on its computational resources according to a dynamic focal point that is important to speech processing with broad hypothesis space, and apply every kind of constraint, from phonetic to pragmatic, at the same time.

## Correspondence between Conversational Situations and Facial Displays

The speech dialogue subsystem recognizes a number of typical conversational situations that are important to dialogues. We associate these situations with an appropriate facial display(s). For example, in situations where speech input is not recognized or where it is syntactically invalid, the listener comment display "Not confident" is displayed. If the speaker's request exceeds the range of the system's domain knowledge, then the system displays a facial shrug and replies "I cannot answer such a question." The relationships between conversational situations and facial displays are listed in Table 2.

## Example Dialogue

Here is an example of a possible dialogue with the system[1]. **Un:** indicates user speech inputs, **Sn:** indicates voice responses delivered by the system to the user. Terms enclosed within square brackets illustrate the types of facial displays introduced in previous section. "Continuing" means that the system facial expression remains the same.

**U1:** Hello.
**S1:** [Attend and BOStory] Hi. This is Sony Computer Science Laboratory.
**S2:** [Continuing] I can answer any question about computer-related products.
**U2:** uh ...
**S3:** [NotConfident] I beg your pardon.
**U3:** I want to know about a personal computer.
**S4:** [Question] Do you want to know about a Sony personal computer?
**U4:** No, I don't.
**S5:** [Shrug] I cannot answer such a question.
**U5:** Please tell me about a Sony personal computer.
**S6:** [BOStory] Sony personal computer "QuarterL" is a standard IBM compatible notebook-style personal computer.
**U6:** What can I do with it?
**S7:** [BOStory] You can use all IBM PC software.
**S8:** [Continuing] For example, you can use a word processor, and a spreadsheet.

---
[1] Actually, the system accepts and speaks only Japanese. The example is a translation.

Table 2: Relation between Conversational Situations and Facial Displays

| CONVERSATIONAL SITUATION | FACIAL DISPLAY(S) |
|---|---|
| Recognition failure | NotConfident (Listener comment display "Not confident") |
| Syntactically invalid utterance | NotConfident |
| Many recognition candidates with close scores | ModConfident (Listener comment display "Moderately confident") |
| Beginning of a dialogue | Attend (Listener comment display "Indication of attendance") |
| Introduction to a topic | BOStory (Syntactic display "Beginning of a story") |
| Shift to another topic | EOStory (Syntactic display "End of a story") and BOStory |
| Clarification dialogue | Question (Syntactic display "Question mark") |
| Underline a remark | Underliner (Syntactic display "Underliner") |
| Answer "Yes" | SpeakerYes (Speaker display "Yes") |
| Answer "No" | SpeakerNo (Speaker display "No") |
| Out of the domain | Shrug (Speaker display "Facial shrug") |
| Answer "Yes" with emphasis | SpeakerYes and Emphasizer (Syntactic display "Emphasizer") |
| Violation of pragmatic constraints | Incredulity (Listener comment display "Incredulity") |
| Reply to "Thanks" | ListenerYes (Listener comment display "Yes") and Smile (Complementary display "Smile") |
| ...... | ......... |

**U7:** Can I use UNIX with it?
**S9:** [BOStory] If you want to use UNIX,
**S10:** [Underliner] I recommend you get a workstation.
**U8:** Tell me about a workstation.
**S11:** [EOStory and BOStory] Sony workstation "NEWS" is a high-performance laptop workstation.
**S12:** [Continuing] Its CPU is an R3081 RISC processor, and its processing speed of 37 MIPS is the fastest in this class.
**U9:** Is it large?
**S10:** [SpeakerNo and Emphasizer] No, it isn't.
**S11:** [BOStory] "NEWS" is 32.4 cm in width, 36.4 cm in depth, and 6.9 cm in height.
**U10:** Is it light?
**S12:** [SpeakerYes and Emphasizer] Yes, it is.
**S13:** [BOStory] The weight of "NEWS" is 4.5 kg.
**U11:** How much?
**S14:** [BOStory] "NEWS" costs 700,000 yen.
**U12:** What does the personal computer cost?
**S15:** [EOStory and BOStory] "QuarterL" costs 398,000 yen.
**U13:** Thank you.
**S16:** [ListenerYes and Smile] You are welcome.
**S17:** [Continuing] It's my pleasure.

U2 is an example of noisy inputs. The system could not recognize the expression and displayed the facial backchannel NotConfident and replied "I beg your pardon." In U3, there is missing information about personal computer maker. So, the system enters a clarification dialogue S4, showing the Question display. In this case, the system tried to drive the user into the domain with which the system is familiar. However, the user refused to agree with the system by utterance U4, then the system revealed its discouragement by showing a facial shrug. In U8, the user changes the topic, by asking for workstation information. The system recognizes this by comparison with the prior topic (i.e., personal computers). Therefore, in response to question S11, the system displays EOStory and subsequently BOStory to indicate the shift to a different topic. The system also manages the topic structure so that it can handle anaphora and ellipsis in utterances such as U9, U10, and U11.

## Experimental Results

To examine the effect of facial displays on the interaction between humans and computers, experiments were performed using the prototype system. The system was tested on 32 volunteer subjects. Two experiments were prepared. In one experiment, called **F**, the subjects held a conversation with the system, which used facial displays to reinforce its response. In the other experiment, called **N**, the subjects held a conversation with the system, which answered using short phrases instead of facial displays. The short phrases were two- or three-word sentences that described the corresponding facial displays. For example, instead of the "Not confident" display, it simply displayed the words "I am not confident." The subjects were divided into two groups, **FN** and **NF**. As the names indicate, the subjects in the **FN** group were first subjected to experiment **F** and then **N**. The subjects in the **NF** group were first subjected to **N** and then **F**. In both experiments, the subjects were assigned the goal of enquiring about the functions and prices of Sony's computer products. In each experiment, the subjects were requested to complete the conversation within 10 minutes. During the experiments, the number of occurrences of each facial display was counted. The conversation content was also evaluated based on how many topics a subject covered intentionally. The degree of task achievement reflects how it is preferable to obtain a greater number of visit more topics, and take the least amount of time possible. According to the frequencies of appeared facial displays and the conversational scores, the conversations that occurred during the experiments can be classified into two types. The first is "smooth conversation" in which the score is relatively high and the displays "Moderately confident," "Beginning of a story," and "Indication of attendance" appear most often. The second is "dull conversation," characterized by a lower score and in which the displays "Neutral" and "Not confident" appear more frequently.

The results are summarized as follows. Details of the experiments were presented in another paper (Takeuchi and Nagao, 1993).

1. The first experiments of the two groups are compared. Conversation using facial displays is clearly more successful (classified as smooth conversation) than that using short phrases. We can therefore conclude that facial displays help conversation in the case of initial contact.

2. The overall results for both groups are compared. Considering that the only difference between the two groups is the order in which the experiments were conducted, we can conclude that early interaction with facial displays contributes to success in the later interaction.

3. The experiments using facial displays **F** and those using short phrases **N** are compared. Contrary to our expectations, the result indicates that facial displays have little influence on successful conversation. This means that the learning effect, occurring over the duration of the experiments, is equal in effect to the facial displays. However, we believe that the effect of the facial displays will overtake the learning effect once the qualities of speech recognition and facial animation have been improved.

The premature settings of the prototype system, and the strict restrictions imposed on the conversation inevitably detract from the potential advantages available from systems using communicative facial displays. We believe that further elaboration of the system will greatly improve the results. The subjects were relatively well-experienced in using computers. Experiments with computer novices should also be done.

# Concluding Remarks and Further Work

Our experiments showed that facial displays are helpful, especially upon first contact with the system. It was also shown that early interaction with facial displays improves subsequent interaction, even though the subsequent interaction does not use facial displays. These results prove quantitatively that interfaces with facial displays help to break down the mental barrier that many users have toward computing systems.

As a future research direction, we plan to integrate more communication channels and modalities. Among these, the prosodic information processing in speech recognition and speech synthesis are of special interest, as well as the recognition of users' gestures and facial displays. Also, further work needs to be done on the design and implementation of the coordination of multiple communication modalities. We believe that such coordination is an emergent phenomenon from the tight interaction between the system and its ever-changing environments (including humans and other interactive systems) by means of situated actions and (more deliberate) cooperative actions. Precise control of multiple coordinated activities is not, therefore, directly implementable. Only constraints or relationships among perception, conversational situations, and action will be implementable.

To date, conversation with computing systems has been over-regulated conversation. This has been made necessary by communication being done through limited channels, making it necessary to avoid information collision in the narrow channels. Multiple channels reduce the necessity for conversational regulation, allowing new styles of conversation to appear. A new style of conversation has smaller granularity, is highly interruptible, and invokes more spontaneous utterances. Such conversation is closer to our daily conversation with families and friends, and this will further increase familiarity with computers.

Co-constructive conversation, that is less constrained by domains or tasks, is one of our future goals. We are extending our conversational model to deal with a new style of human-computer interaction called *social interaction* (Nagao and Takeuchi, 1994) which includes co-constructive conversation. This style of conversation features a group of individuals where, say, those individuals talk about the food they ate together in a restraurant a month ago. There are no special roles (like the chairperson) for the participants to play. They all have the same role. The conversation terminates only once all the participants are satisfied with the conclusion.

We are also interested in developing interactive characters and stories as an application for interactive entertainment. We are now building a conversational, anthropomorphic computer character that we hope will entertain us with some pleasant stories.

## ACKNOWLEDGMENTS

The authors would like to thank Mario Tokoro and colleagues at Sony CSL for their encouragement and helpful advice. We also extend our thanks to Nicole Chovil for her useful comments on a draft of this paper, and Satoru Hayamizu, Katunobu Itou, and Steve Franks for their contributions to the implementation of the prototype system. Special thanks go to Keith Waters for granting permission to access his original animation system.